\documentclass[twoside,11pt]{article}

\usepackage{automl2020}
\usepackage{listings}
\usepackage[frozencache=true,cachedir=minted-cache]{minted}
\usepackage{placeins}
\usepackage{amsmath}
\usepackage{bm}
\usepackage{subcaption}
\usepackage{tikz}
\usepackage{soul}
\usepackage{tikz-qtree}
\usepackage{enumitem}
\usepackage{fancyvrb}
\usepackage{mathtools}
\usepackage{multirow}
\usepackage{algorithm}
\usepackage[noend]{algpseudocode}
\usepackage{mathtools}
\usepackage{booktabs}
\usepackage{wrapfig}
\usepackage{inconsolata}

\jmlrheading{David Eriksson$^*$, Pierce I-Jen Chuang$^*$, Samuel Daulton$^*$, Ahmed Aly, Arun Babu, Akshat Shrivastava, Peng Xia, Shicong Zhao, Ganesh Venkatesh, Maximilian Balandat}

\ShortHeadings{\footnotesize Latency-Aware Neural Architecture Search with Multi-Objective Bayesian Optimization}{Eriksson, Chuang, Daulton, Xia, Shrivastava, Babu, Zhao, Aly, Venkatesh, Balandat}
\firstpageno{1}

\begin{document}

\title{Latency-Aware Neural Architecture Search with Multi-Objective Bayesian Optimization}

\author{\name David Eriksson$^*$ \email deriksson@fb.com \\
    \addr Facebook
    \AND
    \name Pierce I-Jen Chuang$^*$ \email pichuang@fb.com \\
    \addr Facebook
    \AND
    \name Samuel Daulton$^*$ \email sdaulton@fb.com \\
    \addr Facebook
    \AND
    \name Peng Xia \email pengxia@fb.com \\
    \addr Facebook
    \AND
    \name Akshat Shrivastava \email akshats@fb.com \\
    \addr Facebook
    \AND
    \name Arun Babu \email arbabu@fb.com \\
    \addr Facebook
    \AND
    \name Shicong Zhao \email zsc@fb.com \\
    \addr Facebook
    \AND
    \name Ahmed Aly \email ahhegazy@fb.com \\
    \addr Facebook
    \AND
    \name Ganesh Venkatesh \email gven@fb.com \\
    \addr Facebook
    \AND
    \name Maximilian Balandat \email balandat@fb.com \\
    \addr Facebook
    \AND
    \begin{center} \addr \textit{$* =$ Equal contribution.} \end{center}
}

\maketitle

\begin{abstract}%
When tuning the architecture and hyperparameters of large machine learning models for on-device deployment, it is desirable to understand the optimal trade-offs between on-device latency and model accuracy.
In this work, we leverage recent methodological advances in Bayesian optimization over high-dimensional search spaces and multi-objective Bayesian optimization to efficiently explore these trade-offs for a production-scale on-device natural language understanding model at Facebook.
\end{abstract}

\section{Introduction}
Neural architecture search (NAS) aims to provide an automated framework that identifies the optimal architecture for a deep neural network machine learning model given an evaluation criterion such as the model's predictive performance.
The continuing trend towards deploying models on end user devices such as mobile phones has led to increased interest in optimizing multiple competing evaluation criteria to achieve an optimal balance between predictive performance and computational complexity (e.g. total number of flops), memory footprint, and latency of the model.
To address the NAS problem, the research community has developed a wide range of search algorithms, leveraging reinforcement learning (RL)~\citep{zoph2017neural, tan2019mnasnet}, evolutionary search (ES)~\citep{real2019regularized, liu2018hierarchical}, and weight-sharing~\citep{cai2020onceforall, yu2020bignas, wang2021attentivenas}, among others.
However, RL and ES can incur prohibitively high computational costs because they require training and evaluating a large number of architectures.
While weight sharing can achieve better sample complexity, it typically requires deeply integrating the optimization framework into the training and evaluation workflows, making it difficult to generalize to different production use-cases.

In this work, we aim to bridge this gap and provide a NAS methodology that requires \emph{zero} code change to a user's training flow and can thus easily leverage existing large-scale training infrastructure while providing highly sample-efficient optimization of multiple competing objectives.
We employ Bayesian optimization (BO), a popular method for black-box optimization of computationally expensive functions that achieves high sample-efficiency~\citep{frazier2018tutorial}.
BO has been successfully used for tuning machine learning hyperparameters for some time~\citep{snoek2012practical,turner2021bayesian}, but only recently emerged as a promising approach for NAS~\citep{white2019bananas,falkner2018bohb,kandasamy2018neural,shi2019multi,parsa2020bayesian}.

\section{Use-case: On-Device Natural Language Understanding}
\label{sec:model}
We focus on tuning the architecture and hyperparameters of an on-device natural language understanding (NLU) model that is commonly used by conversational agents found in most mobile devices and smart speakers.
The primary objective of the NLU model is to understand the user's semantic expression, and then to convert that expression into a structured decoupled representation that can be understood by downstream programs.
As an example, a user may ask the conversational assistant ``what is the weather in San Francisco?''.
In order for the assistant to reply back with an appropriate answer, e.g., ``The weather in San Francisco is Sunny, with a high of $80$ degrees'', it needs to first understand the user's question.
This is the primary objective of the NLU model, which converts the user's semantic expression into a representation that can be understood by downstream tasks.
We adapt a structured semantic representation of utterance to accomplish this task.
In this setting, the NLU model takes the aforementioned query as input and generates the following structured semantic representation, \,[IN: GET\_WEATHER\; [SL: LOCATION\; San Francisco ] ], where IN and SL represent the \textit{intent} and \textit{slot}, respectively.
The downstream program can then easily identify that the user is interested in the current weather in San Francisco.

\begin{figure}[!ht]
    \centering
    \includegraphics[width=0.5\linewidth]{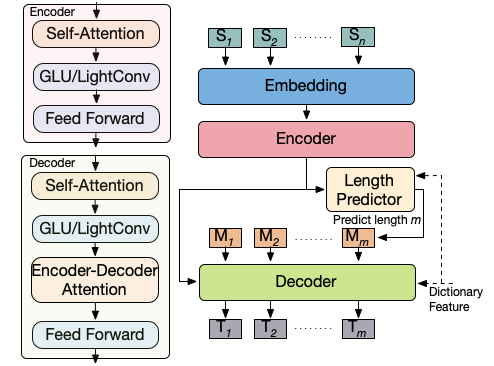}
    \caption{
        Non-Autoregressive Model Architecture of the NLU Semantic Parsing.
    }
    \label{fig:architecture}
\end{figure}

The NLU model shown in Fig.~\ref{fig:architecture} is an encoder-decoder non-autoregressive (NAR) architecture~\citep{babu2021nonautoregressive} based on the span pointer formulation~\citep{shrivastava2021span}.
The span pointer parser is a recently proposed semantic formulation that has shown to achieve state-of-the-art results on several task-oriented semantic parsing datasets~\citep{shrivastava2021span}. An example of the span form representation is shown in Table~\ref{tab:decoupled-comparison}.

\begin{table*}[!ht]
    \centering
    \begin{tabular}{ll}
        \toprule
        Utterance & what is the weather in San Francisco \\
        Index & \;\;\; 1\;\; 2\;\; 3\;\;\;\;\;\;\;\; 4\;\;\;\;\; 5\;\;\; 6\;\;\;\;\;\;\; 7 \\
        \midrule
        Canonical Form & \texttt{[IN: GET\_WEATHER\; [SL: LOCATION\; San Francisco ] ]} \\
        Span Form& \texttt{[IN: GET\_WEATHER\; [SL: LOCATION\; 6 7 ] ]} \\
        \bottomrule
    \end{tabular}
    \caption{Comparison of the canonical and span forms of the decoupled frame representation. Given the utterance ``what is the weather in San Francisco'', our span form produces endpoints instead of text, reformulating the task from text generation to index prediction.}
    \label{tab:decoupled-comparison}
\end{table*}

The NLU model adapts a LightConv architecture consisting of a depth-wise convolution with weight sharing and a transformer-like multi-attention architecture to avoid recurrence.
An input utterance of $n$ tokens (e.g., ``what is the weather in San Francisco?'' has a total of $7$ tokens) first go through the embedding and encoder layer to generate the contextual information.
Then, a convolutional neural network (CNN) based length predictor takes this information and predicts the length of the final output token, labeled as $m$.
For example, [IN: GET\_WEATHER  [SL: LOCATION  San Francisco ] ] requires $8$ output tokens for representation.
On the decoder side, masked tokens with a length $m$ are initialized and propagated through the decoder layer to generate the final output tokens.

As the NLU model serves as the first stage in conversational assistants, it is crucial that it achieves high accuracy as the user experience largely depends on whether the users' semantic expression can be correctly translated.
Conversational assistants operate over the user’s language, potentially in privacy-sensitive situations such as when sending a message.
For this reason, they generally run on the user's device (``on-device''), which comes at the cost of limited computational resources.
While we generally expect a complex NLU model with a large number of parameters to achieve better accuracy, we also want short on-device inference time (latency) so as to provide a pleasant, responsive user experience.
As complex NLU models with high accuracy also tend to have high latency, we are interested in exploring the trade-offs between these two objectives so that we can pick a model that offers an overall positive user experience by balancing quality and delivery speed of the suggestions.

\section{Background}
\subsection{Multi-Objective Optimization}
In multi-objective optimization (MOO), the goal is to maximize\footnote{We assume maximization without loss of generality.}
a vector-valued objective $\bm f(\bm x) \in \mathbb{R}^M$ over a bounded set $ \mathcal X \subset \mathbb{R}^d$.
Typically, there is no single best solution.
Rather, the goal is to identify the Pareto frontier: the set of optimal objective trade-offs such that improving one objective means degrading another.
A point $\bm f(x)$ is \emph{Pareto-optimal} if it is not \emph{Pareto-dominated} by any other point.
With knowledge of the Pareto-optimal trade-offs, a decision-maker can choose an objective trade-off according to their preferences.
Typically, the goal in MOO is to identify a finite, approximate Pareto frontier within some fixed budget.
The quality of a Pareto frontier is commonly assessed according to the \emph{hypervolume} that is dominated by the Pareto frontier and bounded from below by a reference point that bounds the region of interest in objective-space and is usually supplied by the decision-maker.

\subsection{Bayesian Optimization}
\label{sec:bo_and_gps}
Bayesian optimization (BO) is a sample-efficient methodology for optimizing expensive-to-evaluate black-box functions.
BO leverages a probabilistic surrogate model, typically a Gaussian Process (GP)~\citep{Rasmussen2004}.
GPs are flexible non-parametric models that are specified by a mean function $\mu: \mathbb{R}^d \to \mathbb{R}$ and a (positive semi-definite) covariance function $k:\mathbb{R}^d \times \mathbb{R}^d \to \mathbb{R}$.
In this work, we choose the mean function $\mu$ to be constant and the covariance function $k$ to be the popular Mat\'ern-$5/2$ kernel.
In addition to the mean constant, we also learn a signal variance $s^2$, a noise variance $\sigma^2$, and separate lengthscales $\ell_i$ for each input dimension.
We learn the hyperparameters by optimizing the log-marginal likelihood, which has an analytic form~\citep{Rasmussen2004}.

In addition, BO relies on an acquisition function that uses the GP model to provide a utility value for evaluating candidate points on the true black-box functions.
In the MOO setting, common acquisition functions include the expected improvement (EI) with respect to a scalarized objective~\citep{parego}, expected hypervolume improvement (EHVI)~\citep{emmerich2006,daulton2020ehvi}, and information gain with respect to the Pareto frontier~\citep{lobatoa16pesmo,pfes}.

\section{Multi-Objective Bayesian NAS}

We aim to explore the set of Pareto-optimal tradeoffs between model accuracy and latency over a \emph{search space} containing parameters specifying the model architecture from Fig.~\ref{fig:architecture}.

\subsection{Architecture Search Space}
The model architecture hyper-parameters are given in Table~\ref{tab:tunable_parameters} in Sec.~\ref{sup:search_space}.
For each of the main components (encoder, decoder, and the length predictor) in the NLU model, kernel\_list determines the number of layers and the corresponding LightConv kernel size.
As an example, a list of $[3, 5, 7]$ refers to $3$ layers and that the first/second/third layer's LightConv kernel length is $3/5/7$, respectively.
Other important hyperparameters are (1) embed\_dim and ffn\_dim that determine the input and the transformer's feed-forward network (FFN) width, respectively, and (2) attention\_heads for the number of heads in the attention module.
We encode each layer width as an integer parameter and use an additional parameter to control the number of layers included.
This leads to a total of $24$ parameters.

\subsection{Fully Bayesian Inference with SAAS priors}
\begin{figure}[!ht]
\centering
  \includegraphics[width=0.96\linewidth]{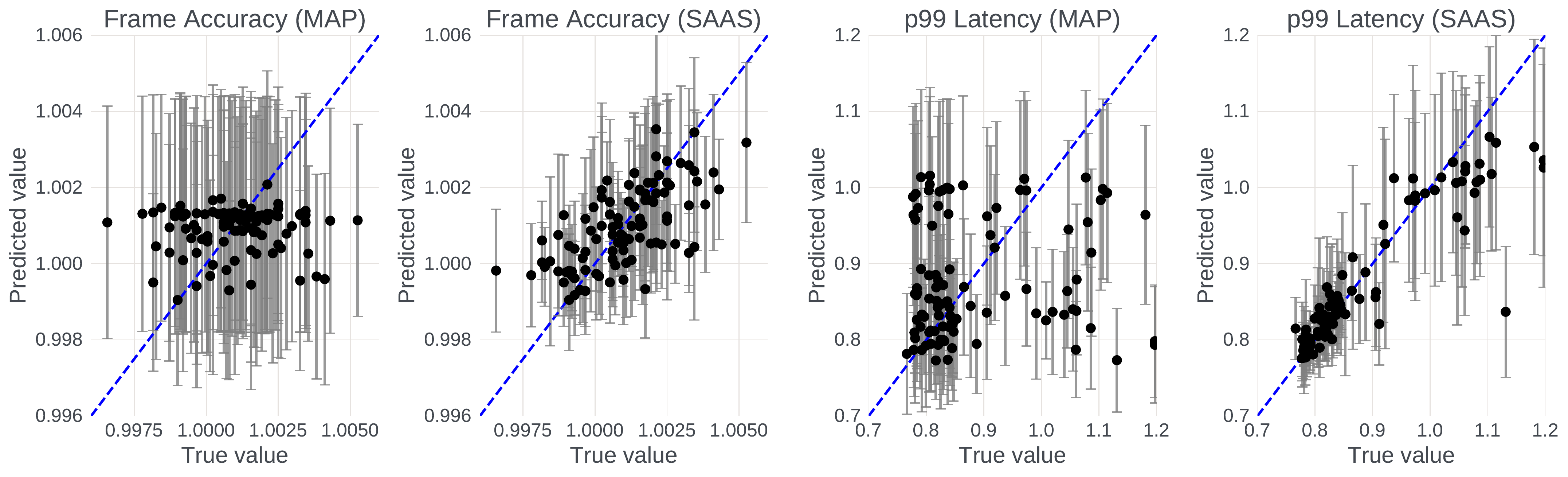}
  \caption{Leave-one-out cross-validation comparison for SAAS and MAP using $100$ training configurations.
  Using the SAAS prior provides good fits for both objectives while MAP estimation is unable to provide accurate model fits.
  }
  \label{fig:saas_vs_map}
\end{figure}

Our 24-dimensional search space poses a challenge for standard GP models as described in Sec.~\ref{sec:bo_and_gps}. In Fig.~\ref{fig:saas_vs_map} we show the leave-one-out cross-validation for two different model fitted to the two objectives for $100$ different input configurations.
We consider objective values relative to the performance of a base model (see Sec.~\ref{sec:experiments} for more details).
We compare a standard GP model with a maximum a posteriori (MAP) approach to the recently introduced sparse axis-aligned subspace (SAAS) prior~\citep{eriksson2021high}.
The SAAS models places sparse priors on the inverse lengthscales in addition to a global shrinkage prior.
When combined with the No-U-Turn sampler (NUTS) for inference, this leads to a model that picks out the most relevant lengthscales, making the model suitable for high-dimensional BO (see Sec.~\ref{sup:gp_fitting} for more details).
As seen in Fig.~\ref{fig:saas_vs_map}, the SAAS prior provides a much better model fit for both objectives, making it a suitable modeling choice.

\subsection{Noisy Expected Hypervolume Improvement}
We use the recently proposed parallel Noisy Expected Hypervolume Improvement acquisition function ($q$NEHVI)~\citep{daulton2021nehvi}, which has been shown to perform well with high levels of parallelism and under noisy observations.
As we use fully Bayesian inference, we integrate the acquisition function over the posterior distribution $p(\psi|\mathcal D)$ of the GP hyperparameters $\psi$ given the observed data $\mathcal D$:
\vspace{-1ex}
\begin{equation}
    \label{eqn:qnehvi_mcmc}
    \alpha_{q\text{NEHVI-MCMC}}(\mathcal X_\text{cand}) = \int_\psi \alpha_{q\text{NEHVI}}(\mathcal X_\text{cand}| \psi)p(\psi|\mathcal D)d\psi
\end{equation}
where $X_\text{cand}$ denotes set of $q$ new candidates $X_\text{cand} = \{\bm x_1, ..., \bm x_q\}$.
Since the integral in~\eqref{eqn:qnehvi_mcmc} is intractable, we approximate the integral using $N_\text{MCMC}$ Monte Carlo (MC) samples (see Sec.~\ref{sup:nehvi} for additional details).

\section{Experimental results}
\label{sec:experiments}
The goal is to maximize the accuracy of the model described in Sec.~\ref{sec:model} on a held-out evaluation set while also minimize the on-device p$99$ latency (the $99$th percentile of the latency distribution).
Latency here defined as the time between when the input utterance becomes available and the model generates the final structured semantic representation.
To produce a stable estimate of the p$99$ latency, a tested model is evaluated repeatedly many times on the same device.
We select a reference point to bound the area of interest based on an existing model with hyperparameters selected using domain knowledge.
We optimize the objectives \emph{relative} to this reference point, which results in a reference point of $(1, 1)$.

\begin{figure}[!ht]
\centering
    \includegraphics[width=0.94\linewidth]{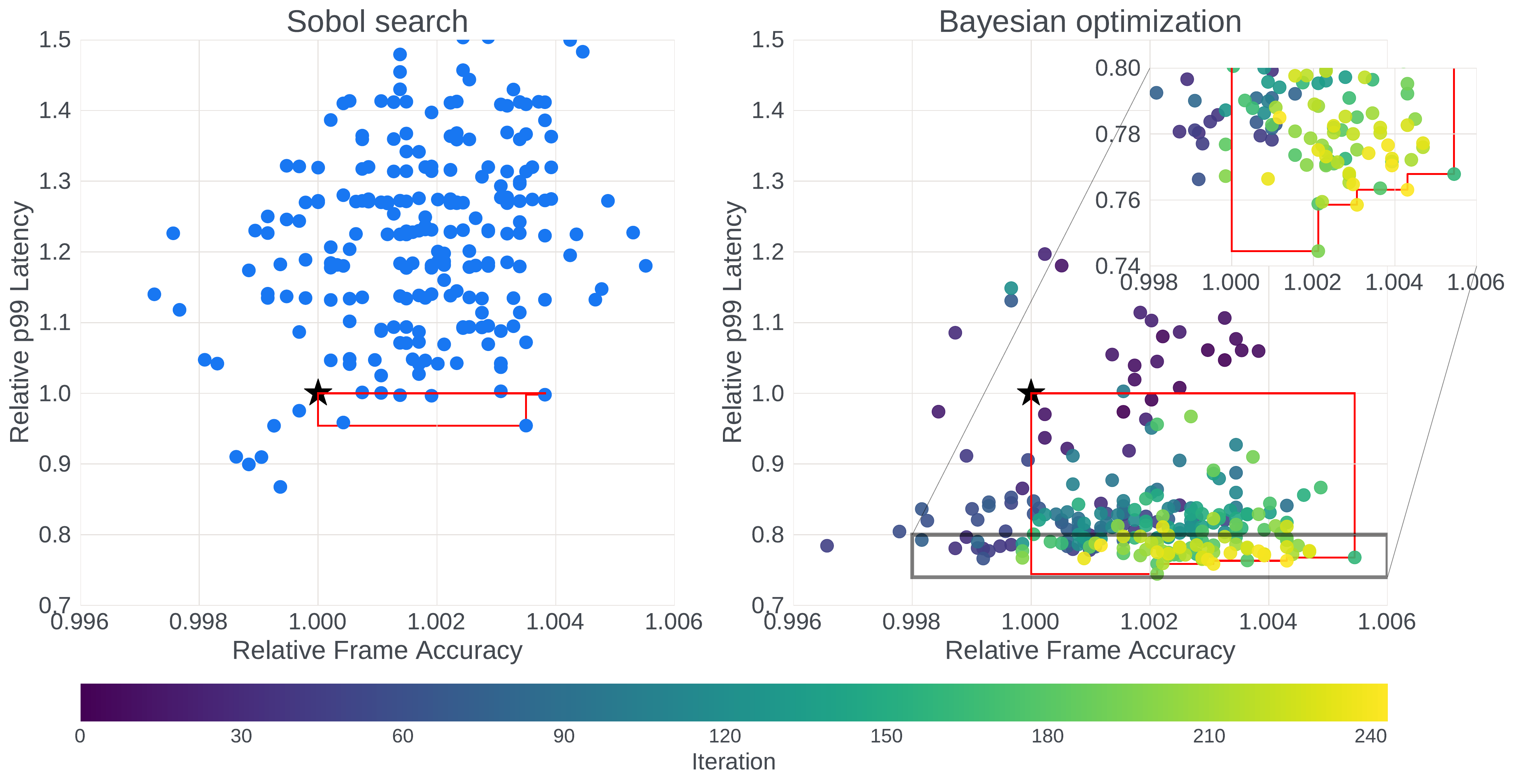}
    \caption{
        (Left) Sobol search is only able to find two points that improve upon the reference point.
        (Right) BO is able to successfully explore the trade-off between latency and accuracy.
    }
    \label{fig:bo_results}
\end{figure}

We consider a computational budget of $240$ evaluations and launch function evaluations asynchronously with parallelism of $q=16$.
For BO, we use $32$ initial points from a scrambled Sobol sequence.
To do inference in the SAAS model we rely on the open-source implementation of NUTS in Pyro~\citep{bingham2019pyro}.
We use $q$NEHVI as implemented in BoTorch~\citep{balandat2020botorch}.
The results from using BO as well as Sobol search are shown in  Fig.~\ref{fig:bo_results}.
Sobol was only able to find a single configuration that outperformed the reference point.
On the other hand, our BO method was able to explore the trade-offs and improve the p$99$ latency by more than $25$\% while also improving model accuracy.

\section{Conclusion}
We introduced a new BO method for sample-efficient multi-objective NAS.
Our approach combines the SAAS prior for high-dimensional BO~\citep{eriksson2021high} with the $q$NEHVI acquisition function~\citep{daulton2021nehvi}.
When applied to a production-scale on-device natural language understanding model, our method was able to successfully explore the trade-off between model accuracy and on-device latency.


\vskip 0.2in
\FloatBarrier
{
    \bibliography{automl2020}

\begin{thebibliography}{27}
\providecommand{\natexlab}[1]{#1}
\providecommand{\url}[1]{\texttt{#1}}
\expandafter\ifx\csname urlstyle\endcsname\relax
  \providecommand{\doi}[1]{doi: #1}\else
  \providecommand{\doi}{doi: \begingroup \urlstyle{rm}\Url}\fi

\bibitem[Babu et~al.(2021)Babu, Shrivastava, Aghajanyan, Aly, Fan, and
  Ghazvininejad]{babu2021nonautoregressive}
Arun Babu, Akshat Shrivastava, Armen Aghajanyan, Ahmed Aly, Angela Fan, and
  Marjan Ghazvininejad.
\newblock Non-autoregressive semantic parsing for compositional task-oriented
  dialog, 2021.

\bibitem[Balandat et~al.(2020)Balandat, Karrer, Jiang, Daulton, Letham, Wilson,
  and Bakshy]{balandat2020botorch}
Maximilian Balandat, Brian Karrer, Daniel~R. Jiang, Samuel Daulton, Benjamin
  Letham, Andrew~Gordon Wilson, and Eytan Bakshy.
\newblock {BoTorch: A Framework for Efficient Monte-Carlo Bayesian
  Optimization}.
\newblock In \emph{Advances in Neural Information Processing Systems 33}, 2020.

\bibitem[Bingham et~al.(2019)Bingham, Chen, Jankowiak, Obermeyer, Pradhan,
  Karaletsos, Singh, Szerlip, Horsfall, and Goodman]{bingham2019pyro}
Eli Bingham, Jonathan~P Chen, Martin Jankowiak, Fritz Obermeyer, Neeraj
  Pradhan, Theofanis Karaletsos, Rohit Singh, Paul Szerlip, Paul Horsfall, and
  Noah~D Goodman.
\newblock Pyro: Deep universal probabilistic programming.
\newblock \emph{The Journal of Machine Learning Research}, 20\penalty0
  (1):\penalty0 973--978, 2019.

\bibitem[Cai et~al.(2020)Cai, Gan, Wang, Zhang, and Han]{cai2020onceforall}
Han Cai, Chuang Gan, Tianzhe Wang, Zhekai Zhang, and Song Han.
\newblock Once-for-all: Train one network and specialize it for efficient
  deployment, 2020.

\bibitem[Daulton et~al.(2020)Daulton, Balandat, and Bakshy]{daulton2020ehvi}
Samuel Daulton, Maximilian Balandat, and Eytan Bakshy.
\newblock Differentiable expected hypervolume improvement for parallel
  multi-objective {B}ayesian optimization.
\newblock In \emph{Advances in Neural Information Processing Systems 33},
  NeurIPS, 2020.

\bibitem[Daulton et~al.(2021)Daulton, Balandat, and Bakshy]{daulton2021nehvi}
Samuel Daulton, Maximilian Balandat, and Eytan Bakshy.
\newblock Parallel {Bayesian} optimization of multiple noisy objectives with
  expected hypervolume improvement.
\newblock \emph{arXiv preprint arXiv:2105.08195}, 2021.

\bibitem[Emmerich et~al.(2006)Emmerich, Giannakoglou, and
  Naujoks]{emmerich2006}
M.~T.~M. Emmerich, K.~C. Giannakoglou, and B.~Naujoks.
\newblock Single- and multiobjective evolutionary optimization assisted by
  gaussian random field metamodels.
\newblock \emph{IEEE Transactions on Evolutionary Computation}, 10\penalty0
  (4):\penalty0 421--439, 2006.

\bibitem[Eriksson and Jankowiak(2021)]{eriksson2021high}
David Eriksson and Martin Jankowiak.
\newblock High-dimensional {Bayesian} optimization with sparse axis-aligned
  subspaces.
\newblock \emph{Conference on Uncertainty in Artificial Intelligence (To
  appear)}, 2021.

\bibitem[Falkner et~al.(2018)Falkner, Klein, and Hutter]{falkner2018bohb}
Stefan Falkner, Aaron Klein, and Frank Hutter.
\newblock Bohb: Robust and efficient hyperparameter optimization at scale.
\newblock In \emph{International Conference on Machine Learning}, pages
  1437--1446. PMLR, 2018.

\bibitem[Frazier(2018)]{frazier2018tutorial}
Peter~I Frazier.
\newblock A tutorial on {Bayesian} optimization.
\newblock \emph{arXiv preprint arXiv:1807.02811}, 2018.

\bibitem[Hernandez-Lobato et~al.(2016)Hernandez-Lobato, Hernandez-Lobato, Shah,
  and Adams]{lobatoa16pesmo}
Daniel Hernandez-Lobato, Jose Hernandez-Lobato, Amar Shah, and Ryan Adams.
\newblock Predictive entropy search for multi-objective bayesian optimization.
\newblock In \emph{Proceedings of The 33rd International Conference on Machine
  Learning}, pages 1492--1501, 2016.

\bibitem[Kandasamy et~al.(2018)Kandasamy, Neiswanger, Schneider, Poczos, and
  Xing]{kandasamy2018neural}
Kirthevasan Kandasamy, Willie Neiswanger, Jeff Schneider, Barnabas Poczos, and
  Eric Xing.
\newblock Neural architecture search with {Bayesian} optimisation and optimal
  transport.
\newblock \emph{arXiv preprint arXiv:1802.07191}, 2018.

\bibitem[{Knowles}(2006)]{parego}
J.~{Knowles}.
\newblock Parego: a hybrid algorithm with on-line landscape approximation for
  expensive multiobjective optimization problems.
\newblock \emph{IEEE Transactions on Evolutionary Computation}, 10\penalty0
  (1):\penalty0 50--66, 2006.

\bibitem[Liu et~al.(2018)Liu, Simonyan, Vinyals, Fernando, and
  Kavukcuoglu]{liu2018hierarchical}
Hanxiao Liu, Karen Simonyan, Oriol Vinyals, Chrisantha Fernando, and Koray
  Kavukcuoglu.
\newblock Hierarchical representations for efficient architecture search, 2018.

\bibitem[Parsa et~al.(2020)Parsa, Mitchell, Schuman, Patton, Potok, and
  Roy]{parsa2020bayesian}
Maryam Parsa, John~P Mitchell, Catherine~D Schuman, Robert~M Patton, Thomas~E
  Potok, and Kaushik Roy.
\newblock Bayesian multi-objective hyperparameter optimization for accurate,
  fast, and efficient neural network accelerator design.
\newblock \emph{Frontiers in neuroscience}, 2020.

\bibitem[Rasmussen(2004)]{Rasmussen2004}
Carl~Edward Rasmussen.
\newblock \emph{Gaussian Processes in Machine Learning}, pages 63--71.
\newblock Springer Berlin Heidelberg, Berlin, Heidelberg, 2004.

\bibitem[Real et~al.(2019)Real, Aggarwal, Huang, and Le]{real2019regularized}
Esteban Real, Alok Aggarwal, Yanping Huang, and Quoc~V Le.
\newblock Regularized evolution for image classifier architecture search, 2019.

\bibitem[Shi et~al.(2019)Shi, Pi, Xu, Li, Kwok, and Zhang]{shi2019multi}
Han Shi, Renjie Pi, Hang Xu, Zhenguo Li, James~T Kwok, and Tong Zhang.
\newblock Multi-objective neural architecture search via predictive network
  performance optimization.
\newblock 2019.

\bibitem[Shrivastava et~al.(2021)Shrivastava, Chuang, Babu, Desai, Arora,
  Zotov, and Aly]{shrivastava2021span}
Akshat Shrivastava, Pierce Chuang, Arun Babu, Shrey Desai, Abhinav Arora,
  Alexander Zotov, and Ahmed Aly.
\newblock Span pointer networks for non-autoregressive task-oriented semantic
  parsing, 2021.

\bibitem[Snoek et~al.(2012)Snoek, Larochelle, and Adams]{snoek2012practical}
Jasper Snoek, Hugo Larochelle, and Ryan~P Adams.
\newblock Practical bayesian optimization of machine learning algorithms.
\newblock In \emph{Advances in neural information processing systems}, pages
  2951--2959, 2012.

\bibitem[Suzuki et~al.(2020)Suzuki, Takeno, Tamura, Shitara, and
  Karasuyama]{pfes}
Shinya Suzuki, Shion Takeno, Tomoyuki Tamura, Kazuki Shitara, and Masayuki
  Karasuyama.
\newblock Multi-objective {B}ayesian optimization using pareto-frontier
  entropy.
\newblock In \emph{Proceedings of the 37th International Conference on Machine
  Learning}, volume 119 of \emph{Proceedings of Machine Learning Research},
  pages 9279--9288. PMLR, 2020.

\bibitem[Tan et~al.(2019)Tan, Chen, Pang, Vasudevan, Sandler, Howard, and
  Le]{tan2019mnasnet}
Mingxing Tan, Bo~Chen, Ruoming Pang, Vijay Vasudevan, Mark Sandler, Andrew
  Howard, and Quoc~V Le.
\newblock Mnasnet: Platform-aware neural architecture search for mobile.
\newblock In \emph{Proceedings of the IEEE/CVF Conference on Computer Vision
  and Pattern Recognition}, pages 2820--2828, 2019.

\bibitem[Turner et~al.(2021)Turner, Eriksson, McCourt, Kiili, Laaksonen, Xu,
  and Guyon]{turner2021bayesian}
Ryan Turner, David Eriksson, Michael McCourt, Juha Kiili, Eero Laaksonen, Zhen
  Xu, and Isabelle Guyon.
\newblock Bayesian optimization is superior to random search for machine
  learning hyperparameter tuning: {Analysis} of the black-box optimization
  challenge 2020.
\newblock \emph{arXiv preprint arXiv:2104.10201}, 2021.

\bibitem[Wang et~al.(2021)Wang, Li, Gong, and Chandra]{wang2021attentivenas}
Dilin Wang, Meng Li, Chengyue Gong, and Vikas Chandra.
\newblock Attentivenas: Improving neural architecture search via attentive
  sampling, 2021.

\bibitem[White et~al.(2019)White, Neiswanger, and Savani]{white2019bananas}
Colin White, Willie Neiswanger, and Yash Savani.
\newblock {BANANAS}: {Bayesian} optimization with neural architectures for
  neural architecture search.
\newblock \emph{arXiv preprint arXiv:1910.11858}, 2019.

\bibitem[Yu et~al.(2020)Yu, Jin, Liu, Bender, Kindermans, Tan, Huang, Song,
  Pang, and Le]{yu2020bignas}
Jiahui Yu, Pengchong Jin, Hanxiao Liu, Gabriel Bender, Pieter-Jan Kindermans,
  Mingxing Tan, Thomas Huang, Xiaodan Song, Ruoming Pang, and Quoc Le.
\newblock Bignas: Scaling up neural architecture search with big single-stage
  models, 2020.

\bibitem[Zoph and Le(2017)]{zoph2017neural}
Barret Zoph and Quoc~V. Le.
\newblock Neural architecture search with reinforcement learning, 2017.

\end{thebibliography}
}


\newpage
\appendix

\section{Example json}
\label{sec:json}
We rely on PyText, which is a deep-learning based language modeling framework built on PyTorch, and is the primary framework for a variety of production NLP models at Facebook, including semantic parsing.
PyText provides a flexible way to generate the deep learning model through a json configuration file, thus providing an easy way to integrate with BO without major code refactoring.
In this simplified example, a $4$-layer encoder, $1$-layer decoder transformer-like model is generated.
The encoder's embedding and feed-forward dimension are $128$ and $192$, respectively.
At single-layer decoder end, the decoder's lightweight convolution's kernel size is $12$, uses GLU for activation, and the attention module is a multi-headed configuration with the number of heads equals to two.
This flexible expression allows the back-end NAS engine to be seamlessly integrated with the model training framework.

\begin{listing}[!ht]
\begin{minted}[frame=single,
               framesep=3mm,
               linenos=false,
               xleftmargin=21pt,
               tabsize=4]{js}
{
    "encoder": {
        "encoder_embed_dim": 128,
        "encoder_ffn_embed_dim": 192,
        "encoder_kernel_list": [3, 5, 5, 7]
    },
    "decoder": {
        "decoder_embed_dim": 156,
        "decoder_output_dim": 128,
        "decoder_glu": true,
        "self_attention_heads": 2,
        "decoder_kernel_size_list": [12]
    }
    // config continues
}
\end{minted}
\caption{Example of transformer-like model json configuration. This type of json configuration is the default in the PyText framework, and consequently any PyText machine learning engineer will be familiar with this format.}
\label{listing:json-example}
\end{listing}

\section{Search space}
\label{sup:search_space}
A summary of the tunable parameters is given in Table~\ref{tab:tunable_parameters}.
The kernel sizes are represented by one integer parameter for each layer and then one additional integer parameter that controls the length.
In particular, we always optimize the acquisition function over all widths as well as the number of layers, but only include the first num\_layers width in the resulting model.
We emphasize that the representation of the kernel sizes could be further improved by taking into account the hierarchical structure, e.g., we do not need a parameter for the width of the $6$th layer if we condition on the encoder length being $5$.
This results in $7$ parameters for the encoder, $3$ for the decoder, and $3$ for the convolution.
In addition, there are $10$ additional integer parameters and $1$ boolean parameter, resulting in a total of 24 parameters.

\begin{table*}[!ht]
\centering
\resizebox{\columnwidth}{!}{%
    \begin{tabular}{lcll}
    Parameter & Default & Search Space & Description \\
    \toprule
    \multicolumn{4}{c}{Encoder} \\
    \midrule
    kernel\_list & [3, 3, 5, 9, 7] & [3, 3, 3, 3], ...  & list of length 4-6, drawn from [3, 5, 7, 9] \\
    embed\_dim & 128 & 128, 136, ..., 192 & input dimension \\
    self\_attention & 2  & 1, 2, 4 & number of self-attention head\\
    ffn\_dim & 40 & 32, 40, ..., 192 & feed-forward network (FFN) width \\
    normalized & True & True, False & apply normalization before the FFN \\
    \midrule
    \multicolumn{4}{c}{Decoder} \\
    \midrule
    kernel\_list & [13, 9] & [7, 7], [7, 9],  ...  & list of length 1-2, drawn from [7, 9, 11, 13, 15] \\
    self\_attention & 1 & 1, 2, 4 & number of self-attention head \\
    attention\_heads & 2 & 1, 2, 4 & number of cross-attention head \\
    ffn\_dim & 144 & 128, 144, ..., 512 & FFN width \\
    \midrule
    \multicolumn{4}{c}{Length Predictor} \\
    \midrule
    kernel\_list & [3, 7] & [3], [3, 5], ...  & list of length 1-2, drawn from [3, 5, 7] \\
    dim & 192 & 32, 40, ..., 192 & convolution width \\
    num\_head & 4 & 1, 2, 4 & number of attention head \\
    \midrule
    \multicolumn{4}{c}{Embedding} \\
    \midrule
    char\_embed\_dim & 8 & 8, 12, ..., 24 & character embedding dimension \\
    proj\_dim & 12 & 8, 12, ..., 24 & last layer projection dimension \\
    \bottomrule
    \end{tabular}
}
\caption{List of the tunable parameters (i.e., search space) of the NLU model}
\label{tab:tunable_parameters}
\end{table*}

\section{GP fitting}
\label{sup:gp_fitting}
Before fitting the GP model we standardize the output values for each objective to have zero mean and unit variance.
We also linearly scale the inputs to lie in the domain $[0, 1]^d$.
Recall from Sec.~\ref{sec:bo_and_gps} that we use a separate lengthscale $\ell_i$ for each input dimension.
Following~\citet{eriksson2021high}, we use a global shrinkage $\tau \sim \mathcal{HC}(0.1)$ and priors $1/\ell_i \sim \mathcal{HC}(\tau)$, where $\mathcal{HC}$ is the half-Cauchy distribution.
As all latent variables in the SAAS model are continuous, we can use the No-U-Turn sampler (NUTS) for inference, integrating out $f$ analytically in the log-marginal likelihood formulation.
The global shrinkage is controlled via $\tau$ and its values will naturally concentrate around zero because of the $\mathcal{HC}$ prior.
As the inverse lengthscales are also governed by a $\mathcal{HC}$ prior, they will also concentrate around zero and the resulting model will, in absence of strong contrary evidence from the data, have large lengthscales and thus ``turn off'' the majority of dimensions.

In addition to the lengthscale priors we use a $\Gamma(0.9, 10)$ prior for the noise variance, a $\Gamma(2.0, 0.15)$ prior for the signal variance, and a $U[-1, 1]$ prior for the constant mean.
For NUTS, we use $512$ warmup steps before collecting a total of $256$ samples.
Finally, we apply thinning and keep only every $16$th sample, leaving us with a total of $32$ hyperparameter samples to average over when computing the acquisition function for any given set of candidate points.

\section{Noisy Expected Hypervolume Improvement}
\label{sup:nehvi}
\citet{snoek2012practical} proposed a similar fully Bayesian treatment of EI, but to our knowledge, no previous work on multi-objective optimization has considered integrating an EHVI-based acquisition function over the posterior distribution of the GP hyperparameters.
Since the integral in~\eqref{eqn:qnehvi_mcmc} is intractable, we approximate it using $N_\text{MCMC}$ Monte Carlo (MC) samples:
\vspace{-1ex}
\begin{equation*}
    \hat{\alpha}_{q\text{NEHVI-MCMC}}(\mathcal X_\text{cand}) = \frac{1}{N_\text{MCMC}}\sum_{n=1}^{N_\text{MCMC}} \alpha_{q\text{NEHVI}}(\mathcal X_\text{cand}| \psi_n)
\end{equation*}
$\alpha_{q\text{NEHVI}}$ also includes an intractable integral and therefore is itself approximated with $N$ MC samples.
We integrate over the posterior distribution of the objectives at pending points to account for the previously selected candidates that are currently being evaluated.
The cached box decomposition approach (CBD) used in $q$NEHVI can be used to efficiently compute $\hat{\alpha}_{q\text{NEHVI-MCMC}}$ by caching $N_\text{MCMC}N$ box decompositions.
See \citet{daulton2021nehvi} for details on efficient computation using CBD.

\section{Hypervolume improvement}
\begin{figure}[!ht]
\centering
  \includegraphics[width=0.55\linewidth]{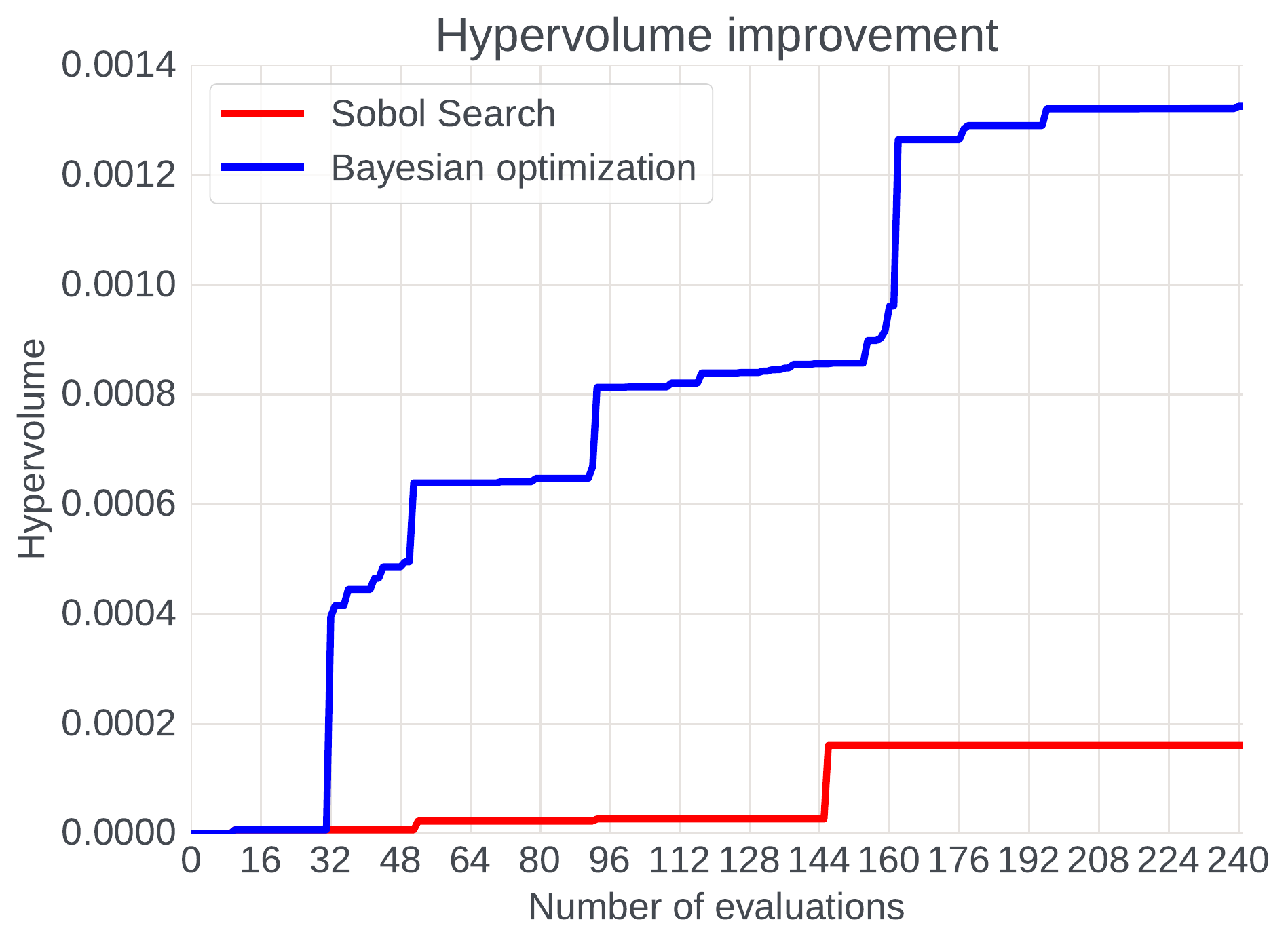}
  \caption{
  We see that BO improves the hypervolume quickly after the initial Sobol batch and makes continuous improvement until the evaluation budget is exhausted.
  }
  \label{fig:hv}
\end{figure}

Fig.~\ref{fig:hv} illustrates that BO achieves a much larger hypervolume compared to Sobol with respect to the reference point $(1, 1)$.
BO immediately makes progress after the initial $32$ Sobol points and consistently makes progress as the GP model becomes more accurate.

\end{document}